\documentclass{article} 
\usepackage{collas2025_conference,times}
\usepackage{easyReview}

\usepackage{amsmath,amsfonts,bm}

\def\eqref#1{equation~\ref{#1}}

\def\1{\bm{1}}

\DeclareMathAlphabet{\mathsfit}{\encodingdefault}{\sfdefault}{m}{sl}
\SetMathAlphabet{\mathsfit}{bold}{\encodingdefault}{\sfdefault}{bx}{n}

\DeclareMathOperator*{\argmin}{arg\,min}

\usepackage{hyperref}
\hypersetup{
    colorlinks=true,
    linkcolor=red,
    filecolor=magenta,
    urlcolor=blue,
    citecolor=purple,
    pdftitle={Overleaf Example},
    pdfpagemode=FullScreen,
    }

\usepackage{graphicx}
\usepackage{booktabs} 
\usepackage{amsmath}
\usepackage{amssymb}
\usepackage{colortbl}
\usepackage{arydshln}
\usepackage{multirow}
\usepackage{algorithm,algorithmic}
\usepackage{amsfonts}
\usepackage{wrapfig}

\title{Warming Up for Zeroth-Order Federated Pre-Training with Low Resource Clients}

\author{Gwen Legate  \thanks{corresponding author gwendolyne.legate@mila.quebec} \\
Concordia University\\
Mila \\
\And 
Irina Rish  \\
University of Montreal\\
Mila \\
\And 
Eugene Belilovsky \\
Concordia University \\
Mila \\
}

\preprintcopy 

\begin{document}

\maketitle
\vspace{-10pt}
\begin{abstract}
\vspace{-5pt}
Federated learning enables collaborative model training across numerous edge devices without requiring participants to share data; however, memory and communication constraints on these edge devices may preclude their participation in training.  We consider a setting in which a subset of edge devices are below a critical memory or communication threshold required to conduct model updates. Under typical federated optimization algorithms, these devices are excluded from training which renders their data inaccessible and increases system induced bias. We are inspired by MeZO, a zeroth-order method used for memory-efficient fine-tuning. The increased variance inherent to zeroth-order gradient approximations has relegated previous zeroth-order optimizers exclusively to the domain of fine tuning; a limitation we seek to correct. We devise a federated, memory-efficient zeroth-order optimizer, $\textbf{ZOWarmUp}$  that permits zeroth-order training from a random initialization. ZOWarmUp leverages differing client capabilities and careful variance reduction techniques to facilitate participation of under-represented, low-resource clients in model training. Like other federated zeroth-order methods, ZOWarmUp eliminates the need for edge devices to transmit their full gradients to the server and instead relies on only a small set of random seeds, rendering the up-link communication cost negligible. We present experiments using various datasets and model architectures to show that ZOWarmUp is a robust algorithm that can can be applied under a wide variety of circumstances. For systems with a high proportion of edge devices that would otherwise be excluded from training, this algorithm provides access to a greater volume and diversity of data, thus improving training outcomes.
\end{abstract}

\section{Introduction}
\vspace{-5pt}
\noindent Over the last decade, there has been a notable increase in data generation, processing, and storage on edge devices. These devices can be used to train advanced neural networks, spanning applications from object detection in autonomous driving to facial recognition on smartphones. Centralized model training methods can become impractical due to bandwidth and memory constraints and the collection and storage of personal data raises concerns about individual privacy rights \citep{zhang2022federated, fan2023model}, particularly in industries where data breaches can lead to serious legal penalties \citep{li2020flchallenges}.  Federated Learning (FL) is a distributed training approach that addresses these challenges by allowing edge devices to train a shared global model collaboratively. It provides a framework where data stored on individual edge devices, also called clients, is never shared. Instead, model training progresses as clients communicate model updates to a coordinating central server \citep{mcmahan2017communication, konevcny2016federated}. FedAvg \citep{mcmahan2017communication} is a popular federated baseline that reduces communication bottlenecks by allowing clients to take multiple gradient steps before communicating with the server. Nevertheless, communications between client and server remain a key bottleneck in FL since inter-node communication costs are orders of magnitude larger than  local communication costs and conducting too many gradients steps prior to communication with the server, may cause client solutions to diverge from the optimal global solution, hindering global model convergence. \citep{zhao2018federated, pmlr-v232-legate23a}. Furthermore, edge devices in practical FL settings can be highly resource heterogeneous. Many clients will have limited compute, network bandwidth and/or memory capacity while others may be capable of rapid communication and vast storage. This can present  a formidable challenge in a federated setting where clients lacking  resources may only be capable of limited participation in federated update rounds or in the worst case, prevent them from contributing altogether \citep{fan2023model}.\\
\vspace{-5pt}

\begin{wrapfigure}{r}{0.5\textwidth}
    \centering
    \includegraphics[width=0.43\textwidth]{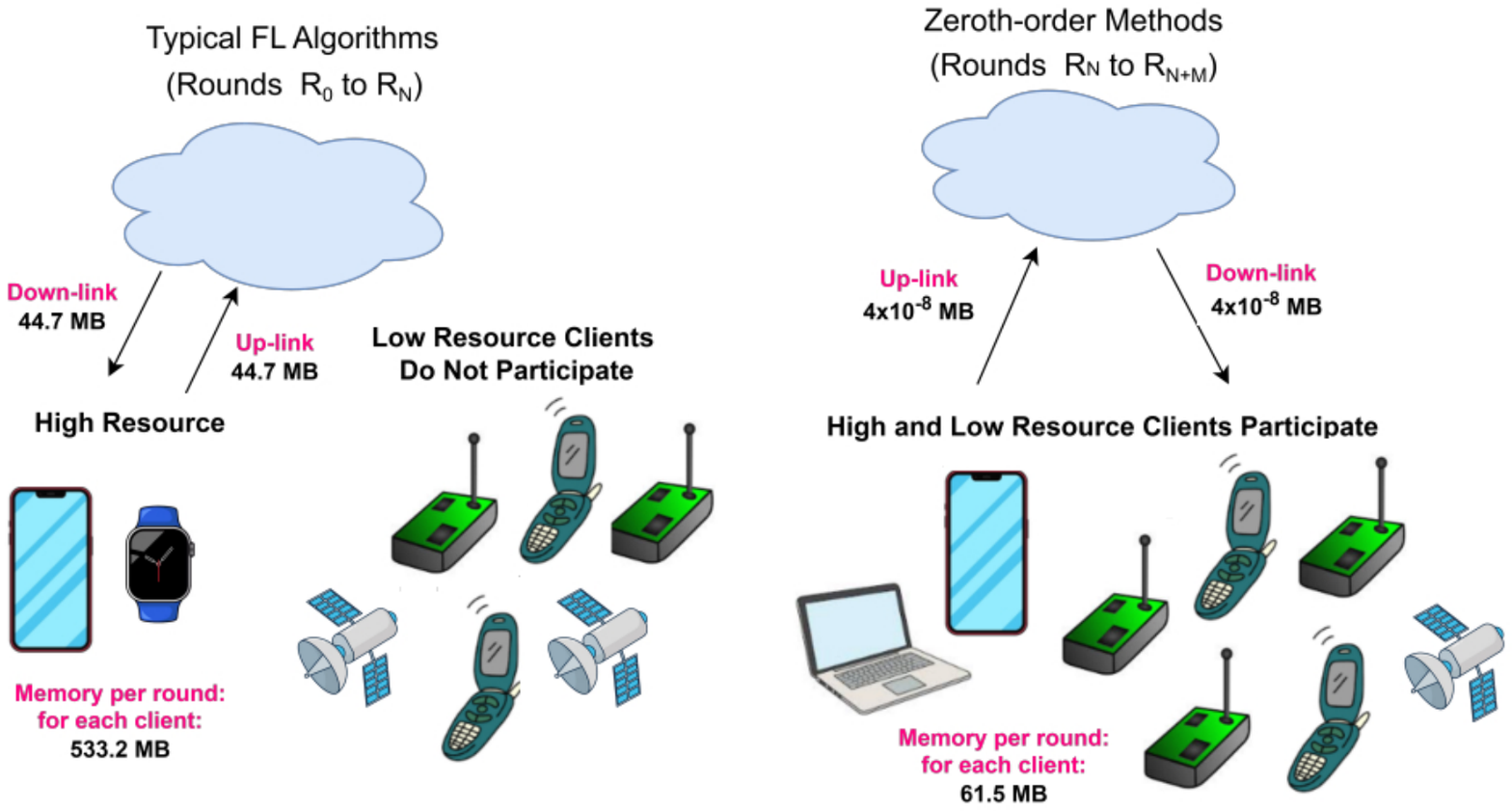}
    \vspace{-5pt}
    \caption{ LHS: Typical federated training where clients with resource constraints are excluded from training. RHS: Zeroth-order training can facilitate low resource client participation but does not achieve good outcomes in pre-training. ZOWarmUp employs a two step method by using high resource training RHS) as a warm-up then performing zeroth-order FL (LHS).}
    \vspace{-8pt}
    \label{fig:resource_footprint}
\end{wrapfigure}

\noindent Resource heterogeneity between clients is an open research problem and several previous publications attempt to address the challenge of fair participation between resource heterogeneous clients \citep{diao2020heterofl, horvath2021fjord, li2023resolving, qiu2022zerofl}. Many approaches focus on client computation constraints, while our work addresses both memory and communication constraints. We consider a setting in which the resource distribution across clients is highly heterogeneous.  Specifically, clients are defined as either high or low resource, where a low resource client is characterized by such severe resource constraints (in terms of memory and communication bandwith) that they are not able contribute to training in any capacity with the model of interest. This setting, depicted in figure~\ref{fig:resource_footprint}, creates system induced bias where the data from clients with better network connections and faster processors will be over-represented during training \citep{kairouz2021advances}. A promising low resource alternative to standard training methods is zero-order optimization \cite{malladi2024fine}. Since it can be achieved using only forward passes, we are not required to store activations; additionally, communication costs can be made insignificant \citep{qin2023federated}. These benefits significantly reduce both memory and communication burdens. 

Zeroth-order optimization relies on  gradient approximations using just forward passes. Since these approximations can possess higher variance than their first-order counterparts, zeroth-order methods such as MeZO \citep{malladi2024fine}, FedKSeed \citep{qin2023federated} and FedZO \citep{fang2022communication} have so far been used exclusively with pre-trained model weights. Extending them to the pre-training setting, common in many federated learning applications, has not been shown in the literature. A challenge is how to combine low resource updates with high resource updates. In this work we devise a two step training regime ($\textit{ZOWarmUp}$) that facilitates training from randomly initialized weights using only data residing on client devices. In this method, high resource clients initially train the model in isolation (step one), providing a warm start for the low resource updates in step two. This shift away from pre-trained model initializations is particularly important for domains with strict data governance and limited data availability. ZOWarmUp is compatible with any zeroth-order optimization method in the literature which gives the user a high degree of flexibility for training purposes.\\

\vspace{-8pt}
\noindent Our contributions in this work are as follows:\\
\vspace{-14pt}
\begin{itemize}
    \item We characterize a specific federated learning setting where high and low resource clients must collaborate to train a model utilizing all their data, instead of excluding low-resource clients from the training process as is typical. We further demonstrate the importance of using all available data during federated training.
    \item We propose and demonstrate the effectiveness of a two step algorithm, \textit{ZOWarmUp} based on zeroth-order methods that demonstrate the first application of zeroth-order methods to federated pre-training. 
    \item We show that a variant of federated zeroth-order optimization that utilizes the Rademacher distribution and takes a single gradient step with the same amount of data outperforms existing variants of federated zeroth-order optimizers developed for federated finetuning
    \vspace{-6pt}
\end{itemize}

\section{Related Work}
\vspace{-5pt}
\subsection{Federated Learning}
\vspace{-5pt}
\noindent Federated learning is a distributed learning setting where a set of decentralized compute nodes, commonly referred to as clients, collaboratively train one model under the guidance of a central server. Client data is never shared between participants, updates are instead communicated to a coordinating central server in the form of model weights or gradients, following a period of local training.\citep{mcmahan2017communication, konevcny2016federated, kairouz2021advances}. FL methods are designed to provide a level of privacy to users by avoiding the communication of data outside of the computing device. Since private information can still be recovered from model weights or gradients, depending on the degree of privacy required, standard FL training techniques often need to be paired with differential privacy and secure aggregation \citep{kairouz2021advances}. Since our work is focused on addressing challenges that arise from resource limitations of edge devices, we follow the well established norms of the literature and  do not consider differential privacy and secure aggregation as they are orthogonal to our work.

\vspace{-6pt}
\paragraph{Low Resource Clients in Federated Learning:}
\noindent In previous work, resource heterogeneity across federated clients is primarily addressed through attempts to reduce the cost of memory and/or communication and by imposing conditions that give clients with fewer resources a fair chance to participate in training. In the domain of model compression, \citet{lin2018deep} propose deep gradient compression (DGC), a method that only communicates gradients with magnitudes exceeding a given threshold. In DGC gradients that do not reach this threshold must be accumulated until the threshold is reached in order to avoid accuracy degradation and ensure model convergence. This accommodation involves gradient storage which creates a significant burden on memory resources. FedSQ \citep{Long2023CommunicationEfficient} employs both sparsity and quantization thereby reducing memory and communication burdens on the client. In order to recover the full model for local training however, clients are still required to store the difference between the original gradient and the compressed gradient locally. As a result local training still bears the full memory burden inherent to the computation of a backwards pass.  Additionally, both sparsity and quantization are known to reduce model performance for high rates of quantization or large pruning fractions. This will further limit the utility of such methods for client nodes with particularly stringent resource limitations. Algorithms encouraging fair client participation include \citet{li2023resolving} whom reduce the burden of communication by decoupling communication and learning via encode/decode operations and \citet{qiu2022zerofl} whom leverage locally sparse weights and activations as well as sparse up-link communications to allow resource constrained clients satisfy their local requirements. HeteroFL \citep{diao2020heterofl}, FjORD \citep{horvath2021fjord}, and FedDrop \citep{wen2022federated} reduce communication costs by creating unique sub-networks for each client based on their resource capabilities. These sub-networks present their own challenges, HeteroFL and FjORD sub-networks are static which results in unequal training of model weights and FedDrop randomly drops at fraction, $p_k$, of neurons in each fully connected layer, excluding other layers that may possess a significant proportion of model weights. Furthermore, \citet{cheng2022does} suggest that FedDrop is detrimental to scaling with better results obtained via simple ensembling methods. While all of these methods achieve memory and/or communication cost reductions, federated zeroth-order optimization methods have negligible communication and memory costs that lower the barrier for participation in an update round; the previously discussed FL methods do not come close to achieving this.
\vspace{-5pt}
\subsection{Zeroth-Order Optimization}
\vspace{-5pt}
\noindent Zeroth-order (ZO) optimization methods approximate gradients using estimates derived from the values of the objective function. In contrast to its first-order (FO) counterpart, ZO optimization does not require gradient computation \citep{liu2020primer, belouze2022optimization} and achieves comparable convergence rates \citep{duchi2015optimal, Nesterov2015RandomGM}. As modern neural networks continue to rapidly increase in size, the memory footprint required to conduct back-propagation can quickly become unsustainable \citep{malladi2024fine, chakrabarti2019backprop}. This is a strong motivator for the use of zeroth-order optimization methods capable of substantially alleviating this burden.

\subsection{Federated zeroth-order Optimization}
\vspace{-5pt}
\noindent Zeroth-order optimization is particularly attractive for federated learning, as edge devices often have limited compute and storage capabilities. Beyond computational efficiency, ZO methods can offer inherent privacy advantages: since clients communicate only a small number of scalar function evaluations, the \textit{global sensitivity} of the mechanism (as per differential privacy) may be reduced as it often depends on the dimensionality of the communicated signal \citep{gratton2022privacy}.
\\

\vspace{-6pt}
\noindent FedZO \citep{fang2022communication} is a ZO optimization algorithm for federated learning. FedZO was designed with black-box federated learning scenarios in mind, where gradient information is not readily available. Due to the nature of these black-box problems, the authors use a zeroth-order method inspired by the zeroth-order augmented Lagrangian method (ZOO-ADMM) \citep{liu2018zo}, and perturb weights according to a uniform distribution over a $d$-dimensional unit sphere. We are interested in more typical machine learning scenarios and therefore we use simultaneous perturbation stochastic approximation (SPSA) \citep{spall1992grad_approx} as a gradient approximator. The most similar ZO federated optimization algorithm to ours is FedKSeed \citep{qin2023federated}. Both of our methods are based on MeZO \citep{malladi2024fine}, use SPSA as a gradient approximator and share similar communication costs and methodologies. These similarities notwithstanding, the key difference between our methods is that ZOWarmUp allows us to tackle training from scratch while FedKSeed focuses on fine tuning LLMs under similar conditions to MeZO. Eliminating the dependence on pre-trained models for ZO optimization makes these methods more broadly applicable since they are no longer tied to a specific domain or model architecture. This task; however, is not trivial and requires careful handling of challenges associated with the reduced precision of approximate gradients.  

\section{Problem Setting and Methods}
\vspace{-5pt}
\noindent We consider a setting in which the edge devices that make up a set of federated clients $K$, have  highly heterogeneous capabilities. A \textit{low resource client} is defined by limitations to one or both of communication or memory so severe that they would not otherwise be capable of contributing to typical federated training; rendering their data inaccessible. Clients are classified as members of the high resource set, $H\subset K$ if they exceed this threshold or the low resource set $L\subset K$ if they do not meet it. The problem setting we describe is highly susceptible to system induced bias, particularly if the proportion of low resource clients is large. 

\noindent In Federated learning, distributed optimization occurs over $K$ clients where each client $k\in\{1, ..., K\}$ possesses data $\mathbf{X}_{k},\mathbf{Y}_{k}$ containing $n_k$ samples drawn from distribution $D_k$. We define the total number of samples across all clients as $n=\sum_{k=1}^{K}n_{k}$. The data $\mathbf{X}_{k}$ at each node may be drawn from different distributions and/or may be unbalanced with some clients possessing more training samples than others. The typical objective function for federated optimization is given by Eq.~\ref{eq:fl_objective} \citep{konevcny2016federated} where $w\in\mathbb{R}^d$ are the model weights and $\mathcal{L}$ is the objective function. 

\vspace{-6pt}
\begin{equation} \label{eq:fl_objective}
\mathbf{w}^*\in \argmin_{\mathbf{w}}\sum_{k=1}^{K}\frac{n_{k}}{n}\mathcal{L}(f(\mathbf{w}, \mathbf{X}_{k}))
\end{equation}

\noindent Simultaneous perturbation stochastic approximation (SPSA) \citep{spall1992grad_approx} is a classical zeroth-order gradient estimator that perturbs all variables simultaneously. In contrast with typical finite difference methods which perturb variables one at a time, SPSA removes the linear computational dependence on network dimension $d$, thereby necessitating only two forward passes per optimization step. For a mini-batch $\mathcal{B}$ that is a subset of a given dataset $(\mathbf{X, Y})$, SPSA approximates the gradient of the objective function using equation~\ref{eq:spsa} where $z\in\mathbb{R}^d$ is sampled randomly from a Gaussian distribution $\mathcal{N}(0, I_d)$. 

\vspace{-6pt}
\begin{equation} \label{eq:spsa}
\hat{\mathbf{g_t}}=\frac{\mathcal{L}(\mathbf{w}+\epsilon \mathbf{z}, \mathcal{B})-\mathcal{L}(\mathbf{w}-\epsilon \mathbf{z}, \mathcal{B})}{2\epsilon}\cdot \mathbf{z}\approx \mathbf{zz^T} \hat{\mathbf{g_t}}
\end{equation}

\noindent Once an approximate gradient has been obtained, ZO-SGD uses the unaltered direction of the gradient estimate as the descent direction and optimizes the parameters according to equation~\ref{eq:zo_update} \citep{malladi2024fine}.

\vspace{-6pt}
\begin{equation}\label{eq:zo_update}
\mathbf{w_{t}}=\mathbf{w_{t-1}}-\eta \hat{g}_t
\end{equation}

\subsection{Communication and Memory Cost Savings of Zeroth-Order Federated Learning} 
\vspace{-5pt}
\label{sec:ZOWarmUp}
\label{sec:resource_footprint}
Zeroth-order methods can permit low resource clients that would not otherwise have the ability to participate in federated training to do so. The memory-efficiency of zeroth-order optimizers in fine-tuning has been leveraged in MeZO \citep{malladi2024fine} and for communication efficiency in fine-tuning in \citet{qin2023federated}. 

Our federated zeroth-order optimization method is described in algorithm~\ref{alg:update_zo}, we note that the method is highly flexible and the second step can be replaced with other variants of zeroth-order methods. To better understand the communication savings consider the second step of algorithm~\ref{alg:update_zo} (\textsc{ZOOpt} and \textsc{ZOUpdate}). The server generates $S$ seeds, to create $S$ distributions for each client, $j$, participating in the round. Seeds are communicated these so that clients can re-generate their unique $\mathbf{z_j}$'s locally.  Clients perturb their local model weights and execute $2n$ forward passes to obtain the scalar values for the two terms in the numerator of equation~\ref{eq:spsa}, $S$ times. If we define $\Delta\mathcal{L}=\mathcal{L}(\mathbf{w}+\epsilon \mathbf{z}, \mathcal{B})-\mathcal{L}(\mathbf{w}-\epsilon \mathbf{z}, \mathcal{B})$, then each client can obtain the $S$ $\Delta\mathcal{L}$s corresponding to their $S$ communicated seeds and communicate only these $S$ floating point numbers back to the server. The server collects these $\Delta\mathcal{L}$s and their associated seeds and communicates the complete list back to the clients so they can calculate all of the required $\hat{g}$ values and aggregate them in a federated adaptation of equation~\ref{eq:zo_update}. Each clients model is then updated according \textsc{ZOUpdate} outlined in algorithm~\ref{alg:update_zo}. This method eliminates the communication of gradient approximations client to server and model weight communication from server to client, providing substantial memory and communication cost savings. 

To see the drastic gains in communication cost we quantify the memory and communication cost of conducting one round of zeroth-order FL when compared to one round of FedAvg in table~\ref{tab:resources}. Values are reported for one client conducting one complete round of training via the specified method using a ResNet18 model ($\approx 1.1\times 10^7$ parameters). Additional details including formulas and specific numerical values used, are provided in section~\ref{sec:appx_com_mem} of the Appendix.

\begin{table*}[b]
\begin{center}
\begin{tabular}{ |c|ccc| } 
&  &  &\\
\toprule
\multirow{2}{*}{Method}&\textsc{Up-link Comm} &\textsc{Down-link Comm}  & \textsc{On-device Mem } \\
  &\textsc{(MB/client)} &\textsc{(MB/client)} &\textsc{ (MB/client)}\\ 
 \midrule
\textsc{FedAvg}&$44.7$ &$44.7$ &$533.2$ \\
\textsc{Zeroth-Order FL}&$(S)\cdot4\times 10^{-6}$&$(SK)\cdot4\times 10^{-6}$ &$89.4$ \\
\bottomrule
\end{tabular}
\end{center}
\vspace{-6pt}
\caption{Up-link and down-link communication cost and memory footprint per client per round for FedAvg and ZO methods. $K$ is the number of clients participating in one update round and $S$ is the number of seeds generated per-client.}
\vspace{-12pt}
\label{tab:resources}
\end{table*}

\subsection{Combating The Variance in Zeroth-Order Optimization} \label{sec:var_reduction}
\begin{figure}[t]
    \centering
    \includegraphics[width=0.68\textwidth]{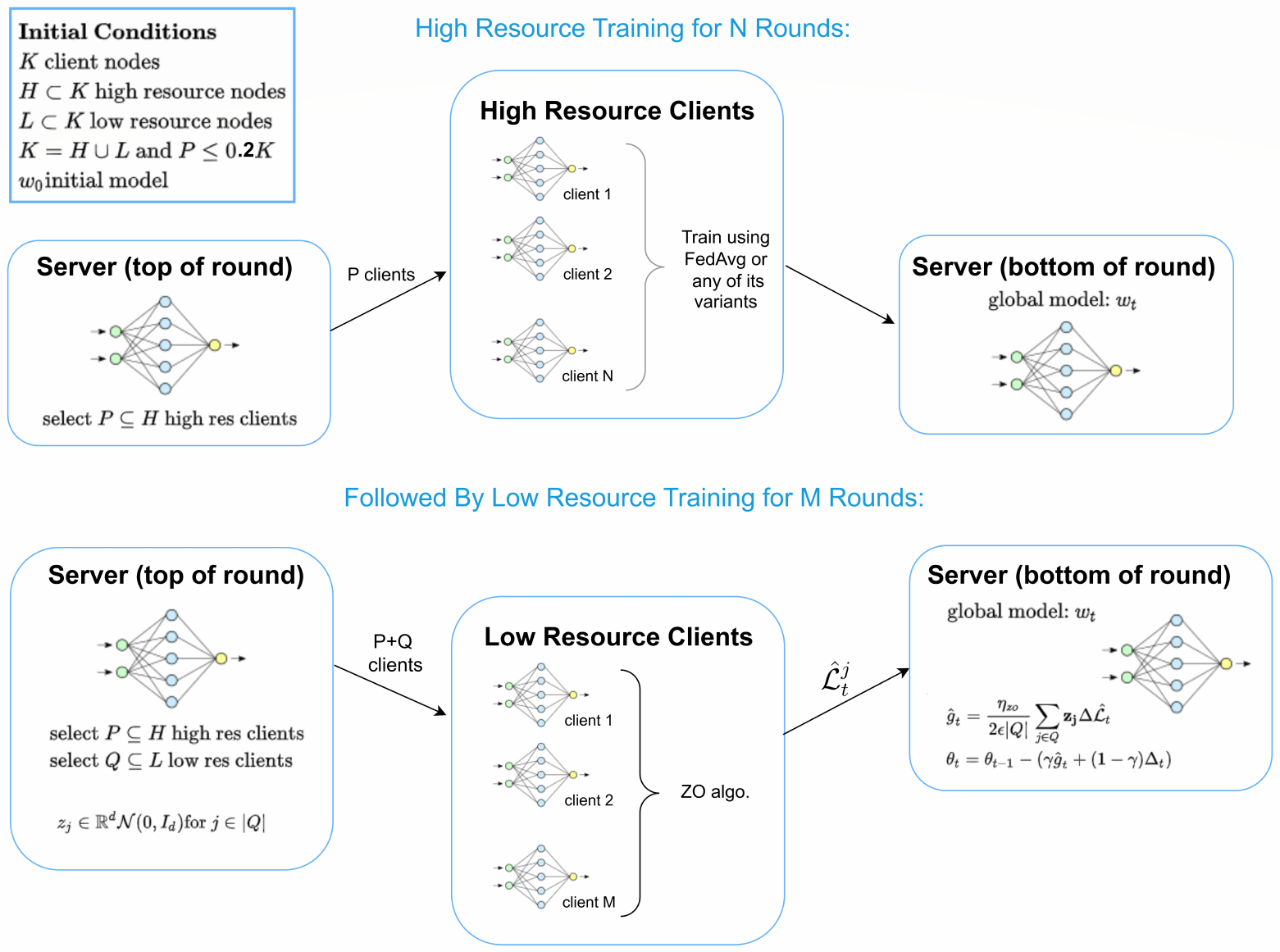}
    \caption{Depiction of ZOWarmUp, a two step process where: \textbf{step 1.)} High resource clients with the capacity to do so train for $N$ rounds using \textit{i.e.}, FedAvg or any of its variants (top). \textbf{step 2.)} Both high and low resource clients make ZO federated updates for an additional $M$ rounds (bottom). }
    \label{fig:method}
    \vspace{-12pt}
\end{figure}
\vspace{-5pt}
\noindent Zeroth-order optimization is considerably noisier than its first order counterpart, this can make it challenging to use when training neural networks from a random initialization. \citet{malladi2024fine} address this problem by using SPSA only to fine-tune pre-trained LLMs whose weights already approach a suitable solution. They additionally introduce hyperparameter $n$, where $n$ is the number of times the model weights are perturbed for each batch. These $n$ perturbations result in $n$ estimated gradients that are subsequently averaged together to give a final gradient estimate and reduce the variance in the update. We make use of this strategy (due to a variable clash, we define \citet{malladi2024fine}'s $n$ variable as $S$ in our work) and additionally adopt the following strategies for variance reduction:

\vspace{-6pt}
\paragraph{Two-Step Training:} The key to training a neural network from scratch using approximated gradients is our two step training regime depicted in figure~\ref{fig:method}. In step one, high resource clients train a model using any of the numerous of federated algorithms existing in the literature. Once updates have stabilized enough to make zeroth-order optimization feasible, we switch completely to low resource updates using a ZO method. This approach works even in scenarios where as little as 10\% of clients have the capacity to participate in high resource training. In subsequent sections we will show the added benefit of accessing previously inaccessible data from these clients cannot be overlooked.

\vspace{-6pt}
\paragraph{Rademacher Distribution for Weight Perturbation:} Work by \citet{belouze2022optimization} suggests that the Rademacher distribution with its lower variance could be a more effective sampling distribution to use with SPSA than the Gaussian distribution. We investigate this possibility in section~\ref{sec:appx_rad} of the Appendix and conclude that the Rademacher distribution does indeed help with convergence in our experiments. We introduce an additional hyperparameter $\tau$, constrained to be $(0, 1)$ that we use to scale the magnitude of the distribution. Please, see section~\ref{sec:appx_exp_hps} in the Appendix for additional experiments regarding both $\tau$ and the previously defined $S$.

\vspace{-6pt}
\paragraph{Single Gradient Step:} In many popular FL algorithms each client typically takes multiple gradient steps before communicating their updates to the central server. This is done to save communication costs but it causes updates to diverge due to the client drift phenomenon \cite{karimireddy2020scaffold}. Since the communication cost of ZO update methods are negligible, we can instead afford to take single gradient steps at each round, while still maintaining drastically lower communication costs than standard FL methods. This strategy has not previously been utilized in any federated ZO optimization algorithms and we demonstrate its advantage by direct comparison to FedKSeed in section~\ref{sec:grad_step}. 

\begin{algorithm}[t]\label{alg:update_zo}
\caption{ZOWarmUp}
\begin{algorithmic}[1]
\REQUIRE $K = H \cup L$ - Total high (H) and low (L) resource clients, $(\mathbf{X}_1,\mathbf{Y}_1), \dots, (\mathbf{X}_K,\mathbf{Y}_K)$ - Local datasets

\STATE \textbf{Initialize:} $\mathbf{w}_0 \forall h \in H$
\FOR{$t = 0, \ldots, N-1$}
    \STATE Sample $P \subseteq H$ high resource clients 
    \FOR{each client $i \in P$, in parallel}
        \STATE $\mathbf{w_{t}^{i}} \gets \textsc{WarmUp}(\mathbf{w_t}, \eta^{c}_{hi})$
    \ENDFOR
    \STATE $\mathbf{w_{t}} \gets \sum_{i=1}^P \frac{|(\mathbf{X}_i,\mathbf{Y}_i)|}{\sum_r^P|(\mathbf{X}_r,\mathbf{Y}_r)|}\mathbf{w_{t}^{i}}$
\ENDFOR
\STATE \textbf{Return} $\mathbf{w}_{N-1}$

\STATE \textbf{Initialize:} $\mathbf{w}_{N-1} \forall \ell \in L$ initialize low res clients with warmed up model weights
\FOR{$t = N, \ldots, N+M-1$}
    \STATE Sample $Q \subseteq K$ high or low resource clients 
    \FOR {each client $j \in Q$, in parallel}
        \STATE $\mathbf{\Delta\mathcal{L}_j}\in \mathbb{R}^S \gets \textsc{ZOOpt}(\mathbf{Seeds_j}, \eta^{c}_{zo})$
    \ENDFOR
    \STATE \textbf{Accumulate} $\mathbf{\Delta\mathcal{L}_{All}} \gets \mathbf{\Delta\mathcal{L}}_j$ for $j\in Q$
    \STATE \textbf{Accumulate} $\mathbf{Seeds_{All}} \gets \mathbf{Seeds}_j$ for $j\in Q$
    \FOR {each client $j \in L$, in parallel}
        \STATE \textsc{ZOUpdate}($\mathbf{Seeds_{All}}, \mathbf{\Delta\mathcal{L}_{All}}$)
    \ENDFOR
\ENDFOR

\vspace{6pt}
{\textsc{ZOOpt}}{($\mathbf{Seeds}, \eta_{zo}^c$)}
    \FOR{s in $\mathbf{Seeds}$}
        \STATE Let $\mathbf{z_s}\in \mathbb{R}^d \gets \tau Rad$ (generated from seed $s$)
            \STATE $\mathbf{\Delta\mathcal{L}_s} = \mathcal{L}(\mathbf{w}_{t-1}+ \mathbf{z_s}\epsilon, (\mathbf{X}_j,\mathbf{Y}_j))-\mathcal{L}(\mathbf{w}_{t-1}- \mathbf{z_s}\epsilon, (\mathbf{X}_j,\mathbf{Y}_j))$ 
    \ENDFOR 
    \STATE \textbf{Return} $\mathbf{\Delta\mathcal{L}}\in \mathbb{R}^S$

\vspace{8pt}

{\textsc{ZOUpdate}}{($\mathbf{Seeds_{All}}, \mathbf{\Delta\mathcal{L}_{All}}$)}
    \FOR{$s, \mathbf{\Delta}\mathcal{L}$ in ($\mathbf{Seeds_{All}}$, $\mathbf{\Delta\mathcal{L}_{All}}$ )}
        \STATE Let $\mathbf{z_s}\in\mathbb{R}^d \gets \tau Rad$ (generated from seed s)
            \STATE $\mathbf{\hat{g}_s} \gets equation~\ref{eq:spsa}$
    \ENDFOR 
    \STATE $\mathbf{\hat{g}}=\sum_s^{|\Delta\mathcal{L}_{All}|}\mathbf{g_s}$
    \STATE $\mathbf{w_{t+1}} \gets \eta^{c}_{zo}\mathbf{\hat{g}}$

\end{algorithmic}
\end{algorithm}


\section{Experiments} \label{sec:experiments}
\vspace{-10pt}

\noindent We use two datasets in our experiments, CIFAR-10 \citep{krizhevsky2009learning} and  ImageNet32 \citep{chrabaszcz2017downsampled}. Training data is partitioned equally between 50 clients using a Dirichlet distribution parameterized by $\alpha=0.1$. In order to simulate the previously defined problem setting where low resource clients are incapable contributing to federated training, clients are randomly assigned as either high or low resource according to a pre-determined resource ratio. In what follows, ratios are denoted as $\%\;high\;resource\;clients\; /\; \%\;low\;resource\;clients$. For example, $10/90$ denotes a $10\%$ high resource, $90\%$ low resource split. The baseline High Res Only shows the effect of excluding low resource clients from training completely. Since clients receive different data distributions, some labels are disproportionately over represented among high resource clients, some are severely under represented and others may not be represented at all. This uneven representation this leads to bias in training outcomes that is more pronounced as the high resource population diminishes. We note that our method outperforms this baseline in all cases, particularly when the proportion of clients is skewed heavily to the low resource category.\\

\vspace{-10pt}
We find optimal learning rates using a grid search method, additional details for specific experiments can be found in section~\ref{sec:appx_exp_hps} in the Appendix. We primarily use a ResNet18 \citep{he2016deep} model for our experiments but provide additional results using ViT \citep{dosovitskiy2021an} in section~\ref{sec:vit}. Unless otherwise specified, the high resource training algorithm is FedAvg \citep{mcmahan2017communication}. High resource training lasts for 200 rounds and clients train for three local epochs prior to communicating with the server. ZO updates last for 300 rounds, with hyperparameters $S=3$ and $\tau=0.75$ all clients switch to low resource training at this point since allowing high resource clients to continue with high resource updates degrades overall performance (see section~\ref{sec:appx_stay_hi} for these experiments). Clients only take one gradient step per update so we set the batch size to be equal to the number of samples at each client. The implications of this choice are studied in greater detail in section~\ref{sec:grad_step}. Our experiments show conclusively that affording low resource clients the opportunity to participate in training, albeit in a more limited capacity, can provide better training outcomes than simply discarding them and their wealth of un-tapped data. Furthermore, our ZO method crafted specifically to reduce variance in gradient estimates outperforms FedKSeed when we use it as the ZO optimization method in the second step of ZOWarmUp.
\vspace{-8pt}

\subsection{ZOWarmUp} \label{sec:main_exps}
\vspace{-5pt}

\begin{table*}[t]
\begin{center}
\scalebox{0.9}{
\vspace{-18pt}
\begin{tabular}{ |c|c|ccccc| } 
\rowcolor{white}
&&&&&&\\
\toprule
&&\multicolumn{5}{|c|}{\textsc{Hi/Lo Split}}\\
\multirow{-2}{*}{\textsc{Dataset}}&\multirow{-2}{*}{\textsc{Method}}&\textsc{10/90}&\textsc{30/70}&\textsc{50/50}&\textsc{70/30}&\textsc{90/10}\\
 \midrule
&HeteroFL&51.1(6.6)&64.5(5.2)&71.3(2.3)&76.2(3.3)&77.1(3.4)\\
 &High Res Only&44.3(6.0)&65.6(3.6)&71.1(1.0)&78.6(0.6)&81.4(1.6)\\
  & FedKSeed &nc&nc&nc&nc&nc\\
  &ZOWarmUp(ours) + FedKSeed&53.4(5.3)&\textbf{73.6(1.6)}&77.3(1.5)&81.2(0.9)&84.4(0.6)\\
 \multirow{-5}{*}{\textsc{CIFAR10}}&ZOWarmUp (ours)&\textbf{54.3(4.8)}&72.4(2.5)&\textbf{77.7(0.5)}&\textbf{82.3(0.8)}&\textbf{84.5(1.3)}\\
 \cdashline{1-7}
 &HeteroFL&\textbf{16.3(0.4)}&20.4(0.8)&20.8(0.7)&20.7(0.8)&19.8(0.8)\\
  &High Res Only&10.9(0.3)&23.7(0.8)&31.5(0.3)&36.4(0.3)&39.9(0.2)\\
 & FedKSeed &nc&nc&nc&nc&nc\\
  &ZOWarmUp(ours) + FedKSeed&11.2(0.7)&24.5(0.4)&\textbf{31.9(0.3)}&37.0(0.2)&40.2(0.3)\\
 \multirow{-5}{*}{\textsc{Imagenet32}}&ZOWarmup(ours)&12.3(0.8)&\textbf{25.3(1.3)}&\textbf{31.9(0.4)}&\textbf{38.8(1.1)}&\textbf{41.5(1.6)}\\
\bottomrule
\end{tabular}
}
\end{center}
\vspace{-2pt}
\caption{Provides the results of our main experiments and presents them in contrast to baselines High Res Only, HeteroFL, and FedKSeed. FedKSeed is evaluated withing the context of our two step ZO training method. Each result is the mean of five seeds $\pm$ the standard deviation. nc indicates not converged}
\label{tab:main}
\end{table*}

\begin{figure}[t]
\centering
\parbox{6.5cm}{
\includegraphics[width=6.5cm]{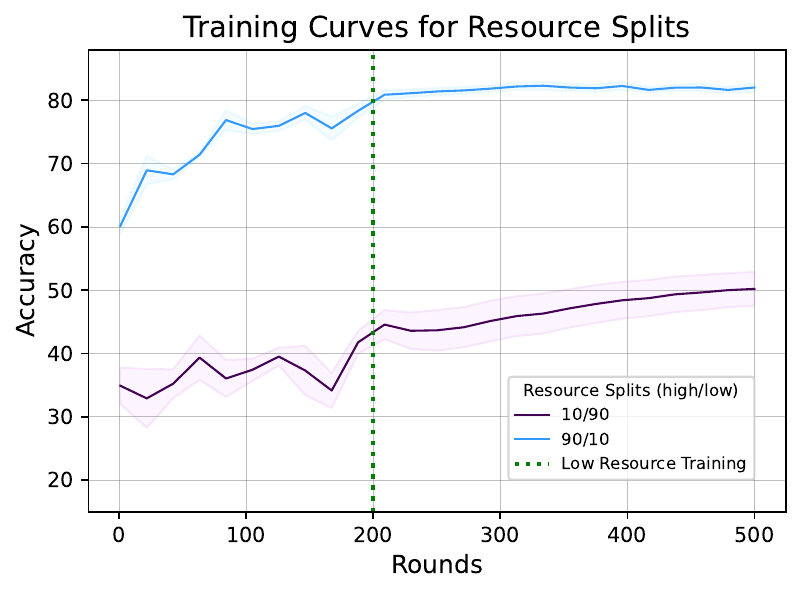}
\vspace{-12pt}
\caption{Training curves for the $10/90$ and $90/10$ resource splits. Both curves show a rapid increase when low resource clients join training at round two-hundred.}
\vspace{-12pt}
\label{fig:curve}}
\qquad
\begin{minipage}{7cm}
\includegraphics[width=6.5cm]{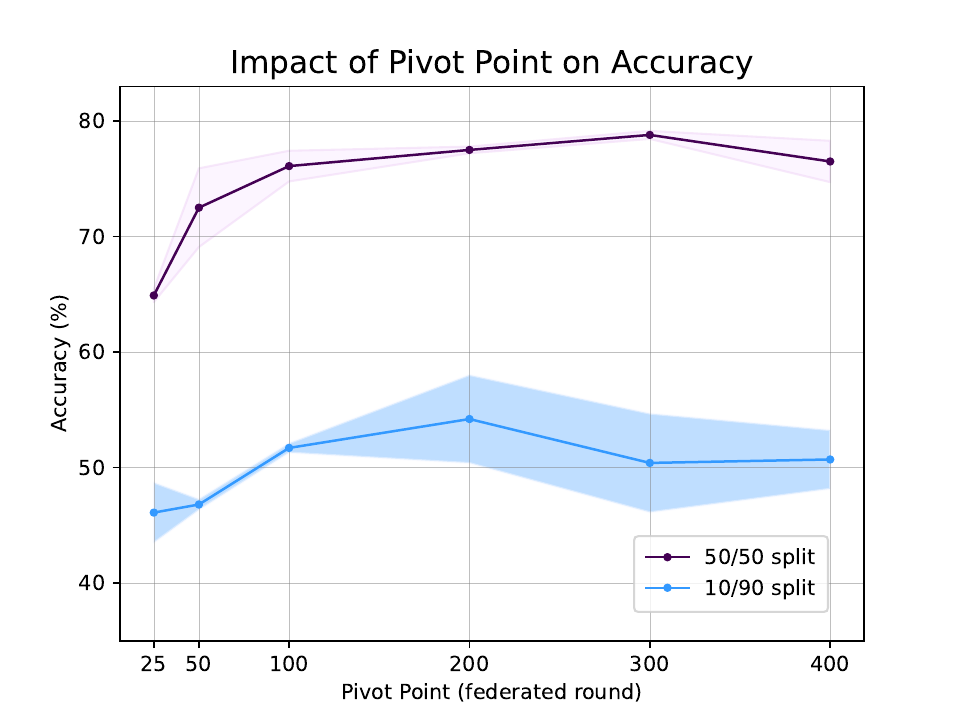}
\vspace{-12pt}
\caption{Shows accuracy as a function of pivot point for two resource distributions. In both cases, a maximum is observed, indicating the need for balance between training too much or too little during step one of ZOWarmUp.}
\vspace{-12pt}
\label{fig:pivot}
\end{minipage}
\end{figure}

Table~\ref{tab:main} compares several baselines; \textit{High Res Only} demonstrates the cost of discarding low resource clients by excluding all clients incapable of participating in training due to their resource constraints. \textit{HeteroFL} \citep{diao2020heterofl}, is a resource efficient FL method that provides clients with a number of sub-networks of varying size based on their resource constraints. We use two HeteroFL sub-networks, high resource clients train using the same ResNet18 we employ in our own experiments and low resource clients train using a ResNet18 that is half the width of the original. We assign a fixed communication budget of the HeteroFL runs that results in less rounds as the proportion of high resource clients increases. \textit{FedKSeed} \citep{qin2023federated} is a ZO optimization method similar to our own; however, it is not able to converge when training from a random weight initialization. In order to provide a meaningful comparison with FedKSeed we employ ZOWarmUp with FedKSeed as the ZO method in step two. This includes only taking a single gradient step for each update to increase the algorithm's stability. The fact that FedKSeed is now able to converge demonstrates the effectiveness of our warm start approach to permit the use of zeroth-order optimization while removing the necessity of a pre-trained model initialization.
\vspace{-2pt}

First we establish that regardless of the ZO algorithm used, continuing to the second step of training is consistently beneficial since \textit{ZOWarmUp + FedKSeed} and \textit{ZOWarmUp} both provide improvement over \textit{High Res Only}. A direct comparison of zeroth-order methods shows that our ZO optimization method comes out ahead for all but one scenario, indicating that we have successfully improved the stability of ZO federated client updates.\\

\vspace{-8pt}
With the exception of the $10/90$ resource split ImageNet32 experiments, ZOWarmUp consistently outperforms HeteroFL. As the fraction of high resource clients increases, the gap between their respective accuracies widens considerably. This discrepancy is particularly noticeable for ImageNet32 where HeteroFL fails improve past the $50/50$ resource split, while, accuracy continues to improve for the ZO methods. Furthermore, after the $50/50$ resource split HeteroFL is no longer within experimental error of the base line including high resource clients. These results suggest the ZO methods have a much greater sensitivity to resource splits which necessitates more careful tuning but offers greater training flexibility than HeteroFL can. As previously mentioned, HeteroFL outperforms the ZO methods when the low resource portion of clients is above $10/90$ for ImageNet32. We note that the use of logit masking during local training employed by HeteroFL is a powerful advantage. A similar method proposed by \citet{pmlr-v232-legate23a} showed conclusively that logit masking during local training, provides consistent improvement for several federated algorithms. We consider that the logit masking advantage combined with the ZO methods greater sensitivity to resource splits is the reason HeteroFL does better under these conditions.

\noindent Figure~\ref{fig:curve} shows select training curves for 10/90 and 90/10 resource splits. The $10/90$ and the $90/10$ curves show similar training patterns, a rougher initial training period using only high resource clients and then a rapid spike in accuracy followed by a smoother period of ZO training. This behaviour is in line with expectations for the 10/90 split, which has access to only a small fraction of high resource clients to train with in the initial phase. Remarkably; however, the 90/10 split which has access to the majority of its client's data during high resource training also sees a visible increase in accuracy when the low resource clients are introduced. This result highlights an important conclusion: \textit{If we have the means to access it, no fraction of data, regardless of how small, should be discarded.} \\  
\vspace{-12pt}

\subsection{The Impact of Gradient Steps} \label{sec:grad_step}
\vspace{-6pt}
In most FL works multi-step training befor aggregation is used to reduce communication cost. However, the utility of such an approach is premised on the idea that the number of samples to process is below a critical batch size. Furthermore, in the case of high heterogeneity the cost of client drift due to increased iteration can be detrimental and gradient accumulation followed by a single step may do better n the same amount of data. We find that in particular for ZO based FL the noisy gradient updates exacerbate the client drift while using the same sample budget to take a single step can reduce the variance and avoid client drift. 

\begin{figure}[t]
    \centering
    \vspace{-12pt}
    \includegraphics[width=0.4\textwidth]{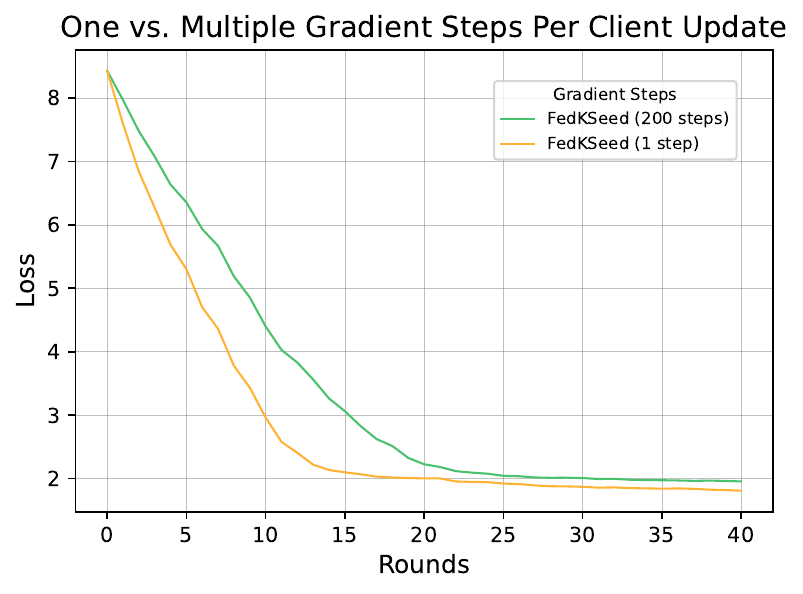}
     \vspace{-5pt}
    \caption{Loss curve comparison of FedKSeed and our proposed one gradient step modification of FedKSeed. Each method uses the same aount of data per round. Final Rouge-L scores are FedKSeep (1 step): $\mathbf{0.2015}$ and FedKSeed (200 steps): $0.1723$.}
    \label{fig:one_step}
    \vspace{-12pt}
\end{figure}

To test this hypothesis, we make a direct comparison with the experimental seedings of existing ZO work found in FedKSeed. We adopt the conditions provided in \citet{qin2023federated} for the DataJuicer-1.3B model \citep{djv1} using the Natural Instructions (NI) \citep{naturalinstructions} dataset and fine-tune the model using FedKSeed (200 gradient steps per client, per round) and our proposed modification that takes only a single gradient step per client on an equivalent aount of data, per round.  Figure~\ref{fig:one_step} shows that over forty rounds of training, taking one gradient step converges faster and reaches a lower loss value than FedKSeed with multiple gradients steps. In keeping with the evaluations done in \citet{qin2023federated}, we use  Rouge-L as the final evaluation metric. Our one step modification of FedKseed achieves a score of $0.2015$, a notable improvement over FedKSeed with a score of 0.1723.  

We now return to ZOWarmUp to study the same effects in our CIFAR-10 and ResNet18  setting. We use an identical high-res training process as described in section~\ref{sec:main_exps} to show the effect of taking multiple gradient steps on performance (see Table~\ref{tab:grad_steps}). Our main experiment conditions where only one gradient step is taken per client, per round has the best training outcomes, regardless of resource split. Larger numbers of gradient steps show pronounced deterioration of in the final model performance. ZO client updates are particularly prone to diverge as client models take independent gradient steps due to the increased variance of the gradient approximations. In addition to deteriorating performance, the hyperparameter $\tau$, responsible for scaling the weight perturbation distribution, needs to decrease drastically to facilitate convergence.  
\vspace{-6pt}

\begin{table*}[t]
\vspace{-28pt}
\begin{center}
\scalebox{0.9}{
\begin{tabular}{ |c|c|ccccc| } 
\rowcolor{white}
&&&&&&\\
\toprule
&&\multicolumn{5}{|c|}{\textsc{Hi/Lo Split}}\\
\multirow{-2}{*}{\textsc{No. Grad Steps (Effective BS)}}&\multirow{-2}{*}{$\tau$}&\textsc{10/90}&\textsc{30/70}&\textsc{50/50}&\textsc{70/30}&\textsc{90/10}\\
 \midrule
0 (900)&0.75&\textbf{54.3(4.8)}&\textbf{72.4(2.5)}&\textbf{77.7(0.5)}&\textbf{82.3(0.8)}&\textbf{84.5(1.3)}\\
1 (450)&0.25&48.3(4.0)&71.2(1.7)&78.0(1.3)&81.2(1.3)&83.4(0.4)\\
4 (255)&0.1&51.6(6.1)&67.9(2.0)&76.8(1.4)&80.1(2.1)&83.1(0.4)\\
6 (150)&0.01&42.2(3.6)&65.4(5.1)&72.7(1.6)&77.4(2.0)&78.5(2.7)\\
\bottomrule
\end{tabular}
}
\end{center}
\vspace{-8pt}
\caption{The effect to increasing gradient steps at the client on ZO optimizer convergence.}
\label{tab:grad_steps}
\end{table*}

\subsection{When to Pivot?}
\vspace{-2pt}
The dotted line in figure~\ref{fig:curve} indicates the switch from high resource training to low resource training at round two-hundred. This transition point which we call the \textit{pivot point} affects training outcomes and must be tuned depending on the task at hand. In this section we set a fixed number of training rounds at five-hundred and ablate the pivot to investigate its effect on training outcomes. Figure~\ref{fig:pivot} shows accuracy as a function of the pivot point for the $10/90$ and the $50/50$ resource splits. For both $10/90$ and $50/50$ there is a sharper increase in accuracy up until a pivot point of 100. The increased amount of data available when $50\%$ of clients are high resource makes the increase in accuracy up to pivot point 100 for the $50/50$ split more impactful than for $10/90$. Interestingly, we observe a maximum at pivot point 200 for the $10/90$ split and at 300 for the $50/50$ split. To explain this phenomenon we reference work by \citet{yan2021critical} who demonstrate the existence of critical learning periods, defined as the first epochs of training, in FL. During the critical learning period, small gradient errors can have an irreversible effect on model training such that no amount of additional training after this period can recover the lost potential. In their experiments, \citet{yan2021critical} withhold a subset of the training dataset on each client until a pre-defined “Recover Round” $M$. After $M$ communication rounds the entire training dataset is recovered and training proceeds as normal. Their results indicate degraded training outcomes as $M$ increases. Our two step training scenario aligns closely to the experimental conditions in \citet{yan2021critical} and the decrease in accuracy we observe after spending too long on high resource training supports the idea that withholding data too long has negative effects on training outcomes. It is important to select the pivot point such that there is a balance between the necessary high resource pre-training required to stabilize the model weights in preparation for ZO updates, and the point at which the effectiveness of introducing low resource client data diminishes. The pivot point will be highly task and architecture dependent and should be considered as a hyperparameter.  
\vspace{-2pt}

\begin{table*}[t]
\vspace{-5pt}
\begin{center}
\scalebox{0.9}{
\begin{tabular}{ |c|c|ccccc| } 
\rowcolor{white}
&&&&&&\\
\toprule
&&\multicolumn{5}{|c|}{\textsc{Hi/Lo Split}}\\
\multirow{-2}{*}{\textsc{Dataset}}&\multirow{-2}{*}{\textsc{Method}}&\textsc{10/90}&\textsc{30/70}&\textsc{50/50}&\textsc{70/30}&\textsc{90/10}\\
 \midrule
 &High Res Only&36.9(1.4)&54.1(4.0)&62.3(0.6)&67.1(0.7)&69.6(1.1)\\
 \multirow{-2}{*}{\textsc{CIFAR10}}&ZOWarmUp (ours)&\textbf{46.4(3.5)}&\textbf{61.1(2.1)}&\textbf{67.0(1.4)}&\textbf{69.7(0.6)}&\textbf{73.3(0.9)}\\
 \cdashline{1-7}
 &High Res Only&5.7(0.6)&8.1(0.4)&8.9(0.2)&9.5(0.4)&10.1(0.5)\\
 \multirow{-2}{*}{\textsc{Imagenet32}}&ZOWarmUp (ours)&\textbf{6.6(0.2)}&\textbf{9.7(0.3)}&\textbf{9.8(0.1)}&\textbf{11.0(0.4)}&\textbf{11.5(0.1)}\\
\bottomrule
\end{tabular}
}
\end{center}
\vspace{-5pt}
\caption{Training results when using FedAdam \citep{reddi2020adaptive} in both high and low resource optimization phases.}
\vspace{-5pt}
\label{tab:adam_results}
\end{table*}

\subsection{FedAdam}
\vspace{-2px}
\noindent In this section we investigate the use of FedAdam \citep{reddi2020adaptive} as the high resource training method in lieu of FedAvg. We also use FedAdam as the server optimizer after we switch to ZO updates in step two of training. These results are presented in table~\ref{tab:adam_results}. For both CIFAR-10 and ImageNet32, we are still able to consistently improve upon the baseline where only high resource clients participate in training. Surprisingly, the FedAvg results presented in table~\ref{tab:main} convincingly outperform FedAdam. While this is not the case in typical federated learning scenarios \citep{reddi2020adaptive}, Adams reliance on the first and second moments of the gradient is problematic in our scenario. Due to the high variance in our gradient approximations, Adams adjustments can be erratic, hindering convergence. Additionally, in the first step of training, the model only has access to a small subset of the training data. Even as the fraction of high resource clients increases we still notice a substantial increase in accuracy in the training curves in figure~\ref{fig:curve} when the low resource client data is introduced, showing that withholding even small amounts of data is detrimental to training. It is therefore reasonable to infer that moment estimates based on gradients from such limited data are unlikely to be representative of the true gradient direction and may cause the optimizer to nudge updates in the wrong direction. This, in turn, provides a less stable initialization for step two, negatively impacting the ZO updates.
\vspace{-2px}

\subsection{ZOWarmUp with Transformers}
\vspace{-2px}

\label{sec:vit}
\begin{table*}[t]
\vspace{-28px}
\begin{center}
\scalebox{0.9}{
\begin{tabular}{ |c|c|ccccc| } 
\rowcolor{white}
&&&&&&\\
\toprule
\vspace{2pt}
&&\multicolumn{5}{|c|}{\textsc{Hi/Lo Split}}\\
\multirow{-2}{*}{\textsc{Dataset}}&\multirow{-2}{*}{\textsc{Method}}&\textsc{10/90}&\textsc{30/70}&\textsc{50/50}&\textsc{70/30}&\textsc{90/10}\\
 \midrule
 &High Res Only&37.8(3.0)&49.9(1.1)&55.0(1.5)&58.3(1.5)&59,3(0.7)\\
         \multirow{-2}{*}{\textsc{CIFAR10}}&ZOWarmUp (ours)&\textbf{42.9(2.9)}&\textbf{52.9(1.1)}&\textbf{57.0(1.4)}&\textbf{59.2(1.1)}&\textbf{60.3(1.2)}\\
\bottomrule
\end{tabular}
}
\end{center}
\vspace{-5px}
\caption{CIFAR-10 training results using the ViT-B/16 \citep{dosovitskiy2021an} model architecture. Each result is the mean of 5 seeds $\pm$ the standard deviation.}
\vspace{-8px}
\label{tab:vit_results}
\end{table*}

Since the release of \citet{vaswani2017attention}, transformers have become ubiquitous in the field of machine learning. While better known for their NLP achievements, they have also achieved impressive results in the field of computer vision \citep{dosovitskiy2021an}. In this section we investigate how ZOWarmUp will respond when applied to transformers using the ViT-B/16 architecture specified in \citet{dosovitskiy2021an}, results are presented in table~\ref{tab:vit_results}. While ZOWarmUp does improve upon the high resource only training baseline, ViT-B/16 under-performs when compared to our main experiments using ResNet18.  This result is not surprising since Vit-B/16 is known to perform worse than ResNet18 when training on smaller datasets \citep{zhu2023understanding}.  
\vspace{-8px}

\section{Conclusion}
\vspace{-4px}
\noindent In FL limited memory and communication constraints on edge devices often restrict their ability to contribute to training. We develop a method, ZOWarmUp, compatible with any ZO optimization method in the literature that allows clients to conduct ZO updates from a random initialization. From there we develop a unique ZO optimization method, that accommodates hardware limitations, particularly when a majority of devices lack the capacity for standard model updates. We are able to show that ZOWarmUp consistently outperforms other resource efficient baselines and demonstrate significant improvements in model accuracy by using zeroth-order optimization to leverage data from a pool of low-resource clients that would have otherwise been sidelined.

\section*{Acknowledgments}We acknowledge support from the Canada Excellence Research Chairs Program, the RBC Borealis Fellowship program, the FRQNT doctoral training scholarship and the FRQNT new scholars grant. This research was made possible with resources provided by Compute Canada, Calcul Quebec and Mila-Quebec AI Institute.

\bibliography{collas2025_conference}
\bibliographystyle{collas2025_conference}
\clearpage
\appendix
\section{Appendix}

\subsection{SPSA With the Rademacher Distribution} \label{sec:appx_rad}
 Using the CIFAR-10 dataset and twelve different seeds whose hyperparameters are identical except for the ZO sampling distribution, we investigate the effect of using the Rademacher distribution instead of the Gaussian distribution as the sampling distribution $z$ for SPSA (equation~\ref{eq:spsa}). Table~\ref{tab:rvsg} shows the average and the standard deviation across the twelve seeds for each distribution. Since the high and low resource clients and their label distributions vary between experiments, we also include the improvement obtained due only to low resource training,  \textit{i.e.}, $\delta_{lo} = f(\mathbf{X}) - f_{hi}(\mathbf{X_{hi}})$ where $f_{hi}(\cdot)$ is the model trained only on the high resource clients and $f(\cdot)$ is the model that has undergone continued training with ZO optimization after the completion of high resource training.\\

\begin{table}[t]
\begin{center}
\begin{tabular}{ |c|cc|cc| } 
\rowcolor{white}
&&&&\\
\toprule
\textsc{Distribution} &\textsc{Acc}&\textsc{Stdv}&\textsc{$\delta_{lo}$}&\textsc{Stdv}\\
 \midrule
$\mathcal{N}(0, 1)$ &49.4&7.7&11.9&2.9\\
Rademacher&65.5&5.2&9.3&1.4\\
\bottomrule
\end{tabular}
\end{center}
\caption{A comparison of variance in the results obtained using both Gaussian and Rademacher distributions to compute the zeroth-order gradient approximations. The Rademacher distribution exhibits considerably lower variance and better overall accuracy than the Gaussian distribution.}
\label{tab:rvsg}
\end{table}

\noindent As shown in Table~\ref{tab:rvsg}, both the final accuracy and $\delta_{lo}$ exhibit lower variance when we use the Rademacher distribution. In our experiments the minimum accuracy obtained when using the Rademacher distribution was 50.4 while for the Gaussian it was 41.3.  This broad variance in results when using the Gaussian distribution caused significantly lower final accuracy than what we observed for Rademacher distribution, we therefore opt to use the Rademacher distribution in lieu of the Gaussian for ZO updates\\
\\

\subsection{Hyperparameters $\tau$ and $S$ for Variance Reduction:} \label{sec:appx_var_hps}

\begin{figure}[t]
    \centering
    \includegraphics[width=0.45\textwidth]{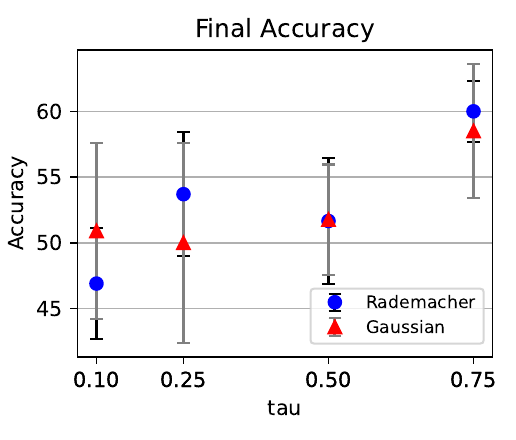}
    \caption{For both plots model pre-trained for 75 rounds using only 10\% of a randomly selected subset of clients designated \textit{high resource}, followed by 425 rounds of training using ZO updates in which all clients participate. The plot shows final accuracy as a function of $\tau$ using both Gaussian and Rademacher distributions. }
    \label{fig:gvsr}
\end{figure}

\paragraph{Scaling with $\tau$:} Figure~\ref{fig:gvsr} shows the effects of scaling the Rademacher and Gaussian distributions by $\tau$. We use $\tau=\{0.75, 0.5, 0.25 0.1\}$, $\tau=1$ is not included since this still permits variance large enough that training can be unstable and not converge for both The Gaussian and Rademecher distributions. Our results indicate a clear advantage to using the Rademacher distribution (scaled by $\tau)$ to approximate the gradients.

\begin{figure}[t]
    \centering
    \includegraphics[width=0.45\textwidth]{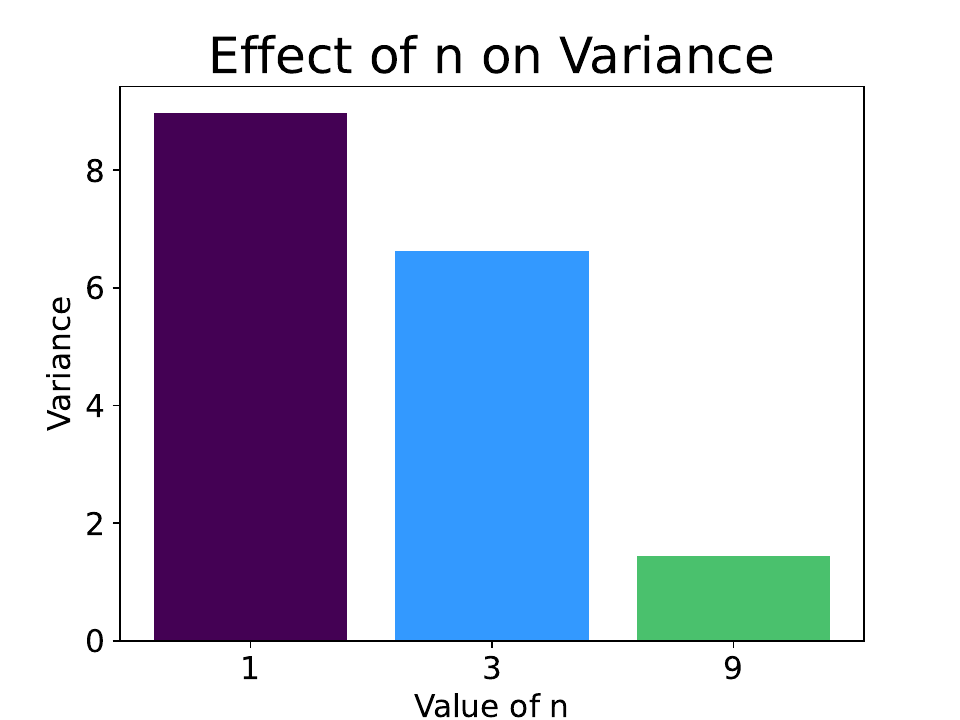}
    \caption{Shows the variance between three seeds for a 10/90 resource split for three values of $S$ . }
    \label{fig:n}
\end{figure}

\paragraph{Sampling $S$ Times:} Figure~\ref{fig:n} demonstrates that increasing $S$ also reduces variance. The improvement between $S=1$ and $S=3$ is 2.4 and between $S=3$ and $S=9$ it is 5.2. Therefore we note that as we continue along the x-axis to greater values of $S$, we observe diminishing returns as $S$ continues to grow. 

\subsection{Communication and Memory} 
\label{sec:appx_com_mem}
\begin{figure*}[t]
    \centering
    \includegraphics[width=0.9\textwidth]{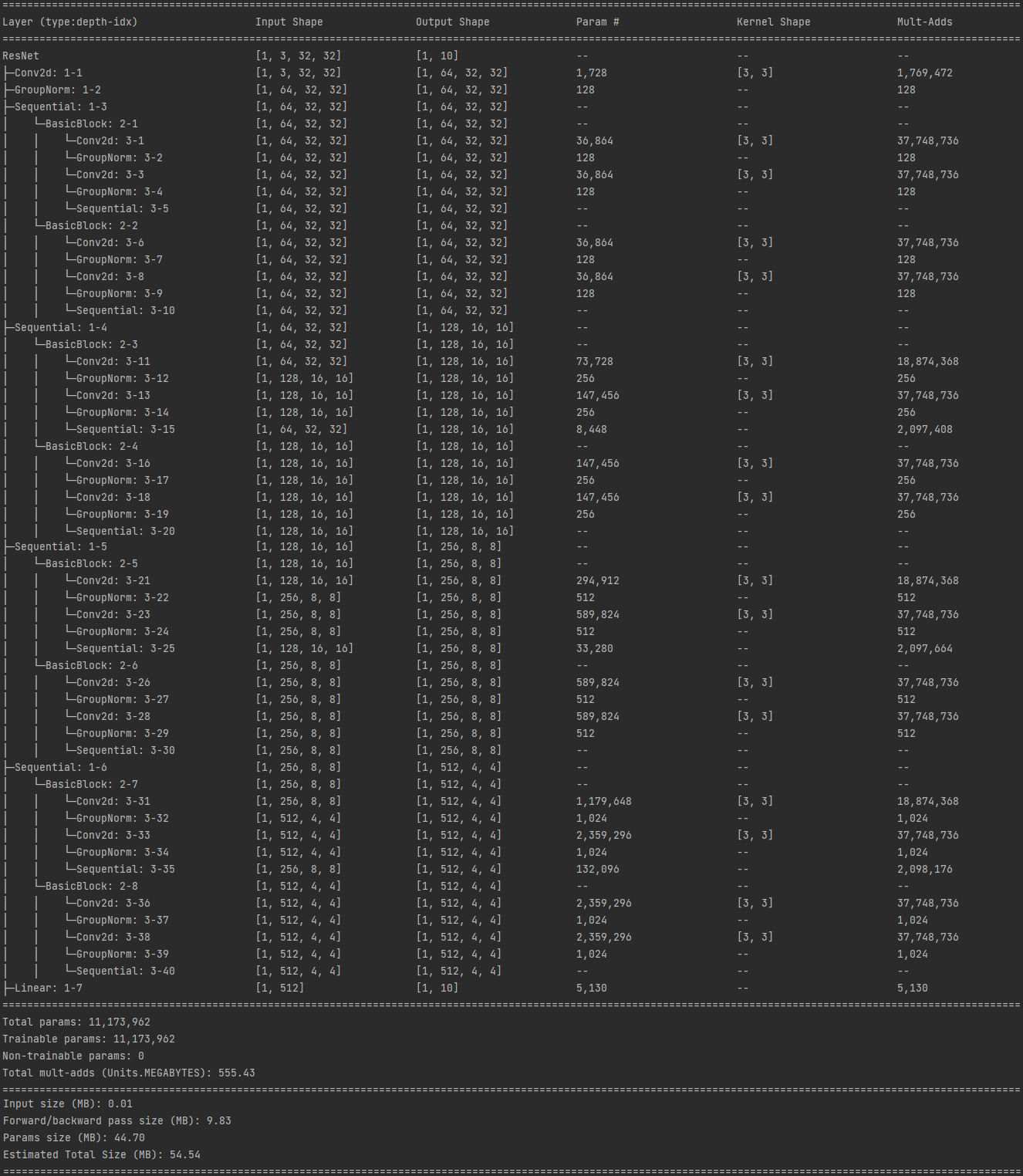}
    \caption{Torchinfo summary of ResNet18 used in our experiments. }
    \label{fig:resnet_stats}
\end{figure*}

\begin{figure*}[t]
    \centering
    \includegraphics[width=0.9\textwidth]{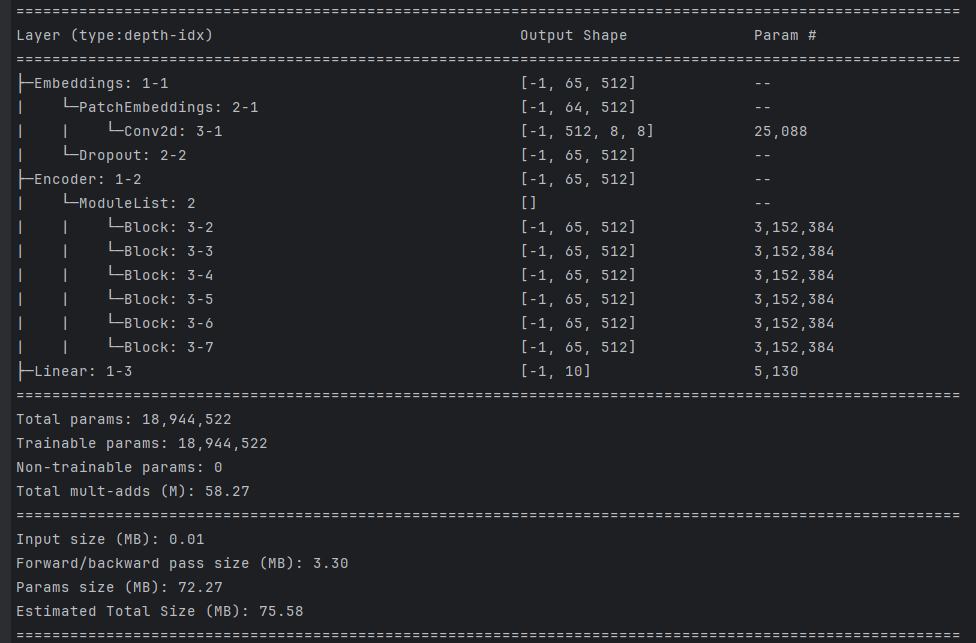}
    \caption{Torchinfo summary of ViT model used in our experiments. }
    \label{fig:vit_stats}
\end{figure*}

Parameter counts and layer dimensions used to calculate memory and communication costs for the different model architectures used were obtained using torchinfo \citep{torchinfo}. Figures~\ref{fig:resnet_stats} and \ref{fig:vit_stats} show the summary provided by torchinfo with the parameter counts used in our calculations.\\

\noindent The following list provides a summary of the notation used in sections \ref{sec:appx_comm} and \ref{sec:appx_mem}:
\vspace{-8pt}

\begin{itemize}
    \item $P$: number of model parameters
    \item $LL_{out}$: output dimension of the classifier
    \item $LL_{in}$: input dimension of the classifier
    \item $\ell\in L$: a model layer
    \item $BS$: batch size
    \item $N_\ell$: number of feature maps in layer output $\ell$
    \item $W_\ell$: width of feature maps in layer output $\ell$
    \item $H_\ell$: height of feature maps in layer output $\ell$
\end{itemize}

\subsubsection{Communication}\label{sec:appx_comm}
For high resource methods a complete set of weights or gradients must be communicated to the server. The communication cost in bits is therefore $comm_{full}=P* (32\;bit)$. ZO methods only need to communicate $S$ floating point numbers per round for both up-link and down-link communications so $comm_{zo}=S*(32\;bit)$. \\

\subsubsection{Memory}\label{sec:appx_mem}
Due to the necessity of storing each output and activation, back-propagation has steep memory costs that grow rapidly with model and/or image size. One backward pass requires storage of the model and the gradient as well as a complete set of model weights and activations. Equation~\ref{eq:mem_full} shows explicitly the additive dependence of a models memory footprint on its outputs and activations.

\small
\vspace{-6pt}
\begin{equation}\label{eq:mem_full}
    mem_{full}=\left( 2P + BS\left(\sum_{\ell\in L} N_\ell* W_\ell* H_\ell \right)\right)* (32\;bit)
\end{equation}

\normalsize
\noindent the steep memory cost of a backward pass makes finding alternatives to back-propagation critical for the inclusion of a greater number of clients in each update. Since weights and activations do not need to be stored if we eliminate the backward pass, we go from an additive dependence on a models outputs and activations to a dependence on only the largest dimension output or activation computed. To provide some empirical context, table~\ref{tab:resources} shows that one round of a ZO optimization method will save $\approx6$ times the memory of one round of FedAvg.

here we provide the explicit formula corresponding to the memory cost of zeroth-order algorithms (equation~\ref{eq:mem_zo}), communication cost is simply $comm_{zero} = n$ since only the seeds must be communicated explicitly.

\small
\begin{equation}\label{eq:mem_zo}
    mem_{zero} = \left( 2P + BS\left(\max\left( N_\ell* W_\ell* H_\ell\right) \right)\right)* (32\;bit)
\end{equation}
\normalsize

\subsection{Combining High and Low Resource Updates}\label{sec:appx_stay_hi}
In this section we evaluate the impact of letting high resource clients continue to make high resource updates, using FedAvg in this case. To make a fair comparison, the data distributions between clients are identical in both cases. Table~\ref{tab:appx_stay_hi} shows the results of these experiments

\begin{table*}[t]
\begin{center}
\scalebox{1}{
\begin{tabular}{ |c|c|ccc| } 
\rowcolor{white}
&&&&\\
\toprule
&&\multicolumn{3}{|c|}{\textsc{Hi/Lo Split}}\\
\multirow{-2}{*}{\textsc{Dataset}}&\multirow{-2}{*}{\textsc{Method}}&\textsc{10/90}&\textsc{50/50}&\textsc{90/10}\\
 \midrule
 &ZOWarmUp(hi+lo)&48.8(5.4)&76.2(1.6)&81.8(0.7)\\
 \multirow{-2}{*}{\textsc{CIFAR10}}&ZOWarmUp(lo only)&\textbf{51.1(4.3)}&\textbf{78.2(0.9)}&\textbf{83.0(0.9)}\\
\bottomrule
\end{tabular}
}
\end{center}
\vspace{-10pt}
\caption{Comparison between having all clients conduct ZO updates and letting high resource clients continue to do high resource updates.}
\vspace{-12pt}
\label{tab:appx_stay_hi}
\end{table*}

\noindent Enforcing ZO updates for all clients regardless of capability provides better training outcomes. We attribute this outcome to the fact that allowing high resource clients to provide more accurate updates may unbalance training and cause greater discrepancies between client updates which will negatively affect convergence.

\subsection{Hyperparameters for Experiments}
\label{sec:appx_exp_hps}
We use two federated algorithms, FedAvg with both client and server optimizer SGD and FedAdam with server optimizer Adam and client optimizer SGD. Unless otherwise specified, experiments train for 500 federated rounds where the first 200 rounds are high resource only \textit{i.e.} FedAdam or FedAvg using only high resource clients, and the last 300 rounds are low resource where all clients participate in updates. The high resource training uses a batch size of 64 and three local epochs. The low resource training training uses one local epoch and a batch size equivalent to the entire client dataset.

\paragraph{ZO Specific Hyperparameters:} As discussed in previous sections, the ZO updates require additional hyperparameters: $\epsilon$ (from equation~\ref{eq:spsa}, $S$ and $\tau$. We consistently set $\epsilon=1\times 10^{-4}$, $S=3$ and $\tau=0.75$. $\tau$ was originally tuned in the range $\{1, 0.75, 0.5, 0.25, 0.1\}$ but we found that $\tau=0.75$ outperformed all other setting regardless of federated algorithm or model architecture. Similar to $\tau$, $\epsilon$ was tuned in the range $\{1\times 10^{-2}, 1\times 10^{-3}, 1\times 10^{-4}, 1\times 10^{-5}\}$ and the $\epsilon=1\times 10^{-4}$ setting consistently outperformed all other options regardless of the setting.

\paragraph{FedAvg:} Hyperparameters for FedAvg include a set of server and client learning rates for both high resource training, $\eta^s$ and $\eta^c$, respectively and low resource training, $\eta^s_{zo}$ and $\eta^c_{zo}$, respectively. They are determined using a grid search over the ranges provided below:\\

\small
\begin{itemize}
\item $\eta^s\in\{1, 0.5, 0.1\}$
\item $\eta^c\in\{5\times 10^{-1}, 1\times 10^{-1}, 5\times 10^{-2}, ..., 5\times 10^{-4}, 1\times 10^{-4}\}$
\item $\eta^s_{zo}\in\{0.5, 0.1, 0.05, 0.01\}$
\item $\eta^c_{zo}\in\{1\times 10^{-1}, 5\times 10^{-2}, 1\times 10^{-2}, ..., 5\times 10^{-6}, 1\times 10^{-6}\}$
\end{itemize}
\normalsize 

\noindent Contrary to conventional practices in federated learning, we find that in our setting using batch normalization with FedAvg outperforms group normalization and we therefore use batch norm for all FedAvg experiments. This advantage does not hold for FedAdam where we revert to group norm.

\paragraph{FedAdam:} FedAdam has additional hyperparameters $\beta 1$ and $\beta 2$ which were set to $0.9$ and $0.999$, respectively. Client and server learning rates for both high resource training were typically lower for Adam, the ranges for these grid searches are provided below:\\

\small
\begin{itemize}
\item $\eta^s\in\{5\times 10^{-3}, 1\times 10^{-3}, ..., 5\times 10^{-6}, 1\times 10^{-6}\}$
\item $\eta^c\in\{1\times 10^{-1}, 5\times 10^{-2}, ..., 5\times 10^{-6}, 1\times 10^{-6}\}$
\item $\eta^s_{zo}\in\{5\times 10^{-3}, 1\times 10^{-3}, ..., 5\times 10^{-6}, 1\times 10^{-6}\}$
\item $\eta^c_{zo}\in\{1\times 10^{-1}, 5\times 10^{-2}, ..., 5\times 10^{-6}, 1\times 10^{-6}\}$
\end{itemize}
\normalsize

\end{document}